\documentclass[10pt,twocolumn,letterpaper]{article}

\usepackage{cvpr}
\usepackage{times}
\usepackage{epsfig}
\usepackage{graphicx}
\usepackage{amsmath}
\usepackage{amssymb}
\usepackage[font+=footnotesize]{subcaption}
\usepackage[font=small]{caption}
\usepackage{enumitem}

\usepackage[pagebackref=true,breaklinks=true,letterpaper=true,colorlinks,bookmarks=false]{hyperref}

\newcommand{\myparagraph}[1]{\smallskip\noindent\textbf{#1.}}
\newcommand{\mysubparagraph}[1]{\smallskip\noindent-- \emph{#1:}}

\cvprfinalcopy % *** Uncomment this line for the final submission

\def\cvprPaperID{1455} % *** Enter the CVPR Paper ID here

\ifcvprfinal\fi
\begin{document}

%%%%%%%%% TITLE
\title{Car Segmentation and Pose Estimation using 3D Object Models}

\author{Siddharth Mahendran and Ren\'e Vidal\\
Center for Imaging Science, Johns Hopkins University, Baltimore MD 21218, USA
}

\maketitle
%\thispagestyle{empty}

%%%%%%%%%%%%%%%%%%%%%%%%%%%%%%%%%%%%%%%%%%%%%%%%%%%%%%%%%%%%%%%%%%%%
% ABSTRACT
\begin{abstract}
Image segmentation and 3D pose estimation are two key cogs in any algorithm for scene understanding. However, state-of-the-art CRF-based models for image segmentation rely mostly on 2D object models to construct top-down high-order potentials.
In this paper, we propose new top-down potentials for image segmentation and pose estimation based on the shape and volume of a 3D object model. We show that these complex top-down potentials can be easily decomposed into standard forms for efficient inference in both the segmentation and pose estimation tasks. Experiments on a car dataset show that knowledge of segmentation helps perform pose estimation better and vice versa.\!\! 
\end{abstract}

%%%%%%%%%%%%%%%%%%%%%%%%%%%%%%%%%%%%%%%%%%%%%%%%%%%%%%%%%%%%%%%%%%%%
% INTRODUCTION
\vspace{-4mm}
\section{Introduction}
\label{sec:introduction}
Given a 2D image, we would like to understand the scene it captures. This involves identifying the objects present in the scene and their spatial extent. This can be done purely in 2D by solving the semantic image segmentation problem (labeling every pixel of the image as belonging to a certain class). However, this is not very informative if say we want to drive an autonomous car without hitting other cars on the road. To a certain extent, true scene understanding requires us to reason about the 3D object shape and scene layout. 

\begin{figure}
	\centering
		\includegraphics[width=0.35\columnwidth]{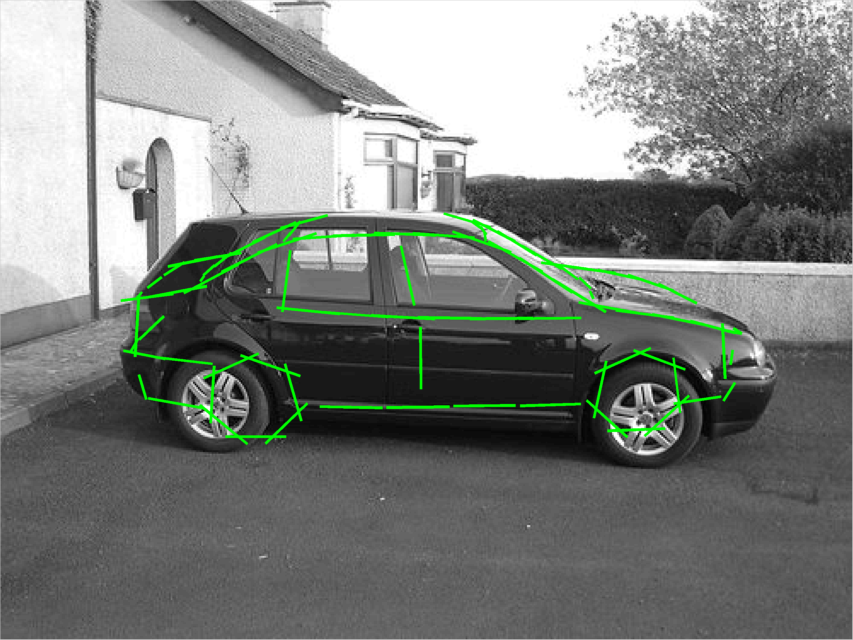}
		\includegraphics[width=0.35\columnwidth]{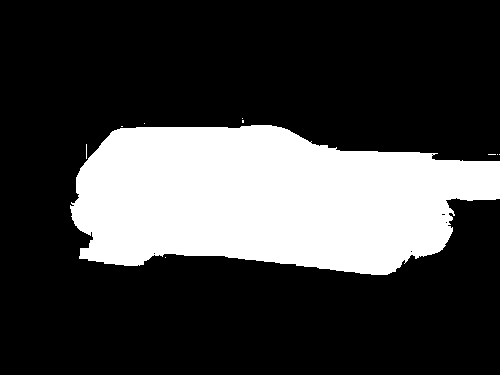}
	\caption{3D model superimposed in image and 2D segmentation.}
	\label{fig:sample-result}
	\centering
	\includegraphics[width=0.7\linewidth,clip=true,trim=40 60 0 0]{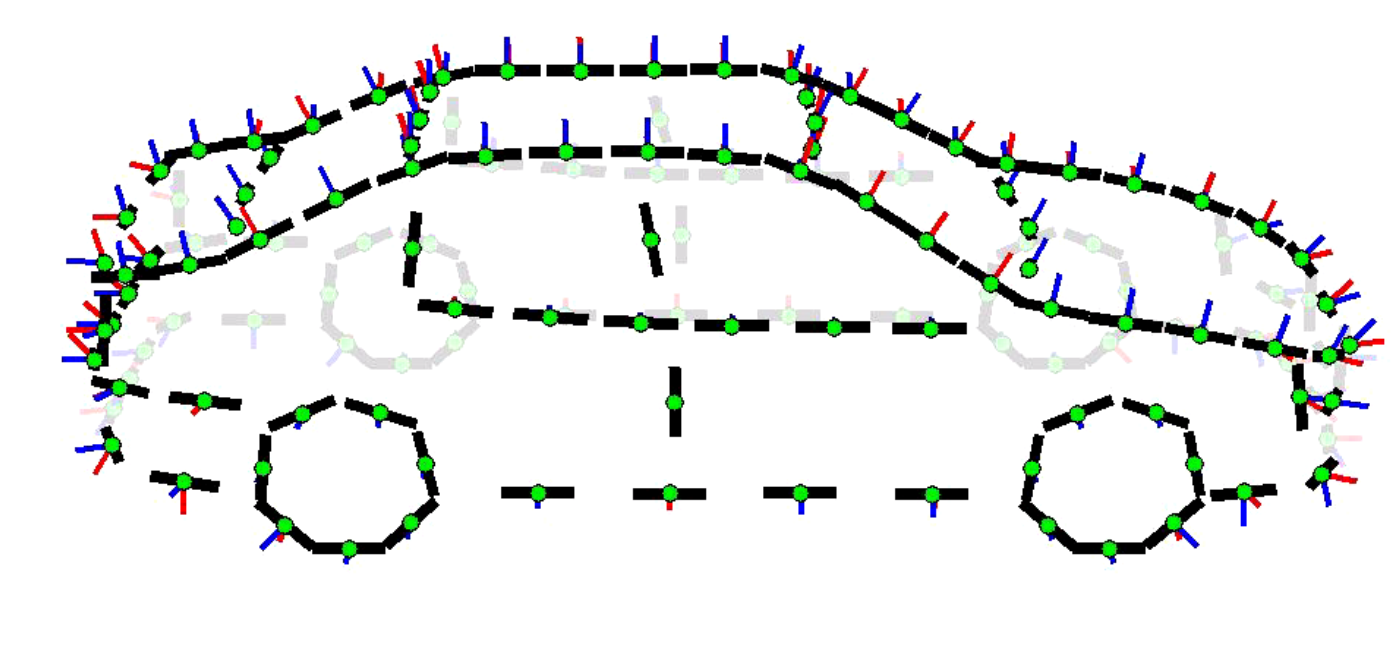}
	\caption{Wireframe model of the class car (taken from \cite{Yoruk:3DRR13}). A single edge primitive consists of the point location (in green), edge direction (in black) and two surface normals (in red and blue).}
	\label{fig:models-wireframe}
	\vspace{-4mm}
\end{figure}

In this paper, we present a small step toward solving this extremely challenging problem. Given a single image, such as that shown in the left of Figure \ref{fig:sample-result}, we concentrate on the problem of segmenting the car (car vs. background) and estimating its 3D pose. Instead of treating 2D segmentation and 3D pose estimation as individual problems, we develop new CRF models for solving both tasks. Our first contribution is to introduce new CRF potentials for semantic segmentation that depend on a 3D object category model. In particular, we use the 3D wireframe model introduced in \cite{Yoruk:3DRR13}, which generalizes 2D HOG features \cite{Dalal:CVPR05} to 3D. Specifically, the wireframe model is a collection of 3D points, edges, and surface normals that, when rotated, translated, and projected, resemble a 2D HOG template. Given a wireframe model, we construct two top-down categorization potentials, called respectively the shape and volume costs, which measure how well:
\begin{enumerate}[topsep=2pt,itemsep=-1ex,partopsep=1ex,parsep=1ex]
\item The projected wireframe model matches the HOG features inside the segment corresponding to the object.
\item The projection of the volume occupied by the wireframe model matches the area occupied by the object in the image (and similarly for the background).
\end{enumerate}

Notice that both costs depend on the 2D segmentation and the 3D object pose. Therefore, the joint 2D segmentation and 3D object pose estimation tasks require the minimization of this energy (properly augmented with standard bottom-up segmentation costs). The energy minimization problem is a challenging mixed continuous and discrete optimization problem, where the energy is non-convex and non-smooth with respect to the pose variables. Our second contribution is to propose an algorithm for solving this problem. In particular, when the object pose is fixed, we show that the top-down costs can be decomposed into standard unary and pairwise terms allowing for efficient image segmentation. Conversely, when the segmentation is fixed, we show that the pose estimation problem, while non-convex and non-smooth, can be efficiently solved using a Branch and Bound (B\&B) algorithm. In particular, we exploit the geometry of the pose space to derive efficiently computable bounds for the shape and volume costs. While the joint segmentation and pose estimation problem could, in principle, proceed in an alternating minimization manner, a key challenge is that B\&B does not produce an exact pose estimate, but rather an optimal pose cell. Therefore, if one chooses an arbitrary pose in the optimal cell, the cost function is not guaranteed to decrease and alternating minimization might fail. We address this issue through a simple pose cell refinement procedure, which guarantees a decrease of the cost. Figure \ref{fig:sample-result} shows the output of our method for one image, where the 3D model is superimposed in the image to visualize its pose. Our experiments on a car dataset will show how segmentation information can be used to improve pose estimation and vice versa.

\section{Related Work} 
\label{sec:related-work}
3D pose estimation and image segmentation are very old problems, which have been studied in great detail. We thus restrict our review to the most relevant literature. 

% ---------------------------------------------- Image segmentation
% using pixels
The CRF model for image segmentation proposed in \cite{Shotton:CVPR08, ShottonJ2006-ECCV} is defined on a pixel graph and uses costs based on shape, texture, color, location and edge cues. 
% top-down using pixels
\cite{Ladicky:ICCV09, Russell:ECCV10} extended the CRF approach to include long-range interactions between non-adjacent nodes in the graph by using costs defined over larger cliques and in a hierarchical manner.
% using superpixels\cite{Ren:ICCV03}
\cite{Fulkerson:ICCV09} extended the CRF approach to a superpixel graph, where an energy consisting of unary and pairwise potentials captures appearance over superpixel neighbourhoods and local interactions between adjacent superpixels.
% top-down using superpixels
\cite{Singaraju:CVPR11,Jain:ECCV12} also introduced new top-down costs based on the bag-of-features classifier \cite{Csurka:ECCV04}, which captures interactions between all regions of an image that have the same label.

%----------------------------------------------- Segmentation using Object models
However, while the top-down costs described above involve large cliques, they are not truly global in the sense that they do not reason at the object or image level (much larger cliques). New top-down costs based on 2D object models can be used instead like in \cite{Yao:CVPR12, Ladicky:ECCV10-detector, Singaraju:ACCV12, Fidler:CVPR13, Dong:ECCV14, Lazebnik:CVPR13}. All these works use the output of object detectors like Deformable Part Models (DPM) \cite{Felzenszwalb:CVPR08,voc-release5} or Exemplar SVM \cite{Malisiewicz:ICCV11} as part of the energy function. Some \cite{Gould:NIPS09, Dong:ECCV14,Singaraju:ACCV12, Yao:CVPR12} also use 2D shape priors to capture object information. These capture longer interactions between different parts of the image and reason at the object level. However, the object models/shape priors used are 2D and hence viewpoint-dependent. %This can be corrected by using 3D object models instead.

%------------------------------------------------ 3D models used for Pose estimation
3D object models have been used for fine pose estimation and have shown to improve 2D detection also significantly. They are mainly of three kinds: obtained by combining 2D object detectors \cite{Schneiderman:CVPR04, Thomas:CVPR06}, extending 2D object detectors to 3D \cite{Savarese-FeiFei:ICCV07, Savarese-FeiFei:ECCV08} or explicit 3D representations \cite{Hu-Zhu:CVPR10, Liebelt-Schmid:CVPR10,Yoruk:3DRR13}. However very few algorithms, like level-set based methods \cite{Sandhu:CVPR09, Dambreville:ECCV08}, use 3D object models for image segmentation. 3D models are very commonly used for segmenting 3D point clouds (robotics) and medical images. But our approach differs from these in that we formulate both pose estimation and image segmentation as energy minimization problems rather than geometric, registration based problems.

%%%%%%%%%%%%%%%%%%%%%%%%%%%%%%%%%%%%%%%%%%%%%%%%%%%%%%%%%%%%%%%%%%%%
% MODELS/ENERGY FUNCTIONS FOR JOINT POSE ESTIMATION AND IMAGE SEGMENTATION
\section{CRF Models for 3D Pose Estimation and 2D Image Segmentation}
\label{sec:models}

A CRF-based approach to image segmentation involves defining a graph $\mathcal{G} = (\mathcal{V},\mathcal{E})$ over the image and an energy on the graph. The image segmentation is then obtained as the labelling that minimizes the energy. We often define $\mathcal{V}$ as the set of all superpixels in the image and $\mathcal{E}$ indicates an edge between two superpixels that share a boundary. A standard energy function defined on this graph is given by 
\begin{equation}
\!\!	E^{u+p}(X;I) \!=\! w_1 \!\sum_{i \in \mathcal{V}} \phi_i(x_i;I) + w_2 \!\!\sum_{(i,j) \in \mathcal{E}} \!\!\psi_{ij}(x_i,x_j;I).\!\!
	\label{eqn:u+p}
\end{equation}
The unary term $\phi_i(x_i;I)$ captures the likelihood of a superpixel being assigned a class label, \ie, $\phi_i(x_i;I)$ is the cost of assigning label $x_i$ to superpixel $i$. These are usually appearance based and are often the score of a classifier applied to features extracted from the superpixel. Hence, they predict the class label of a superpixel without using any additional information. The pairwise term $\psi_{ij}(x_i,x_j;I)$ is a smoothing term, which penalizes assigning different labels to two adjacent superpixels that have a similar appearance. %For sake of conveniece, we define $U(X) = \sum_{i \in \mathcal{V}} \phi_i(x_i;I)$ and $P(X) = \sum_{(i,j) \in \mathcal{E}} \psi_{ij}(x_i,x_j;I)$.

The main disadvantage of using $E^{u+p}$ for segmentation is that the unary and pairwise terms are local in nature as they capture only the interactions between adjacent superpixels. To capture higher order information and long-range interactions between different parts of an object, we often need top-down object categorization costs. In \S \ref{sec:shape-cost} and \S\ref{sec:volume-cost}, we present two new top-down costs that capture such long range interactions using 3D object models. The first one is a \emph{shape cost} that measures how well the projection of the object's 3D shape matches its 2D appearance, and the second is a \emph{volume cost} that measures how well the projection of the object's 3D volume matches its 2D segmentation. Before presenting these new top-down costs, we need to discuss the 3D object model used to construct them. 

%-----------------------------------------------Wireframe model
\subsection{The Wireframe Model}
\label{sec:models-wireframe}
The 3D wireframe model (shown in Figure \ref{fig:models-wireframe}) was introduced in \cite{Yoruk:3DRR13} for the tasks of 2D object detection and 3D pose estimation. The model is a generalization of the 2D HOG features \cite{Dalal:CVPR05} to 3D in which the object is represented as a set of 3D edge primitives $\mathcal{M} = \left \{ p^m, e^m, n_1^m, n_2^m \right \}_{m=1}^M$, where $p^m$ is the location of the primitive in 3D in an object coordinate system, $e^m$ is the 3D edge direction at that point, and $n_1^m$ and $n_2^m$ are the two surface normals at point $p^m$. In \cite{Yoruk:3DRR13}, a car wireframe model was learned from car blueprints in 4 canonical directions (top, front, back and side). Given the model, the car's 3D pose in an image is estimated by maximizing a score that measures how well the projection of the model at a given pose matches the extracted 2D HOG features. The maximization is done using B\&B %a B\&B technique, which will be discussed later in \S\ref{sec:inference-pose_given_seg}.

More specifically, the score $f(T,z;I)$ for estimating the pose $T$ (rotation and translation of the object with respect to the camera) of an object of class $z$ in image $I$ is given by 
\begin{align}
\label{eqn:original-det-cost}
f^{det}(T,z;I) = \sum_{m=1}^{M_z} \nu(p^m_z, n_{1_z}^m,n_{2_z}^m|T) \mathcal{H}(r^m(T,z)).
%&\mathcal{H}(u(p^m_z|T), v(p^m_z|T), w(p^m_z, e^m_z|T)). \nonumber 
\end{align}
Here, $M_z$ is the total number of primitives in a model for class $z$. $\nu(p^m_z, n_{1_z}^m,n_{2_z}^m|T)\in\{0,1\}$ measures whether the 3D primitive is visible or not. A primitive is considered visible in the image if at least one of its surface normals is opposite to the viewing direction. $\mathcal{H}$ are normalized quantized HOG features \cite{Yoruk:3DRR13} extracted from image $I$, where $r^m(T,z) = (u(p^m_z|T), v(p^m_z|T), w(p^m_z, e^m_z|T))$ is the index of the HOG bin to which the $m$-th primitive is projected under pose $T$. The spatial index $(u,v)$ depends only on the 3D location $p_z^m$ but the orientation index $w$ depends also on the 3D edge direction $e_z^m$. The score simply projects each 3D edge primitive onto the image and adds the HOG score at the projected location if that primitive is visible.

%----------------------------------------------- U+P+ Detection 
\subsection{Shape Cost}
\label{sec:shape-cost}

We extend the wireframe model to use segmentation information. We do this by defining a {\it shape score} that is identical to the matching score defined in Eqn.~\eqref{eqn:original-det-cost}, except that now we use only HOG features extracted from regions in the image that belong to the desired class. Let $s^m(T,z)$ be the index of the superpixel containing the projected location of the $m$-th 3D primitive of model $z$ under pose $T$. Note that $r$ and $s$ are related such that, if $r = (u,v,w)$, then $s$ is the superpixel containing $(u, v)$. The shape score is defined as
\begin{equation}
\!\!
f^{seg}(X,T,z;I) = \sum_{m=1}^{M_z} \bar{\mathcal{H}}(r^m(T,z)) \delta(x_{s^m(T,z)}=z),
\label{eqn:modified-det-cost}
\end{equation}
where we write $\bar{\mathcal{H}} = \nu \mathcal{H}$ from Eqn.~\eqref{eqn:original-det-cost} to simplify notation and the $\delta$-Dirac term takes the segmentation into account. 
%Some abuse of notation as the projected point $s$ can be in hog-space (like in the $\bar{\mathcal{H}}$ term) or in pixel-space (like in the $\delta$ term).

The shape score in Eqn.~\eqref{eqn:modified-det-cost} is combined with the standard CRF energy defined in Eqn.~\eqref{eqn:u+p} to get the {\bf shape model} 
\begin{align}
%E^{shape}(X,T,Z;I) &= w_1 U(X) + w_2 P(X) \nonumber \\ &- w_3 f^{seg} (X,T,Z)
E^{shape}(X,T,z;I) = E^{u+p}(X) - w_3 f^{seg} (X,T,z).
\label{eqn:shape-model}
\end{align}
The additional shape cost (negative of shape score) aims to minimize the energy with a higher matching score between the edges in the model and those in the image.

%--------------------------------------------- U+P+ Det. + Comp.+  Residual
\subsection{Volume Cost}
\label{sec:volume-cost}
The shape cost measures how well the 3D edges of the wireframe model match 2D HOG features. However, it is often not enough to just match the edges/boundaries and we also need to match the interior regions. This can be done by measuring the compatibility of the segmentation $X$ with the 2D segmentation induced by the 3D volume corresponding to the pose of the object. For example, we can measure the area of their intersection, $|X \cap \pi(T,z)|$, where $\pi(T,z)$ is a hull of the projection of the wireframe model of class $z$ at pose $T$.\footnote{In what follows we simply use $\pi$ to indicate $\pi(T,z)$ for easy reading.} Notice, however, that the intersection alone is not a good metric as it can be maximized by either assigning all superpixels to the foreground, or choosing the object's pose arbitrarily close to the camera so that the 3D model projects to the whole image.
%{\color{red}Another motivation for this term is to replicate the procedure that is often done in the pose estimation literature. In a pipeline approach, you can find the pose estimate for a given 3D model and then project the model onto the image to generate a segmentation. The term $|X \cap \pi|$ captures this explicitly as part of the energy. Note that it affects only the foreground and if left as is, it tries to capture as much of the segmentation as possible instead of just the car.} 
To address this issue, we also include the term $|X^c \cap \pi^c|$, which measures the area of the intersection for the background. The {\bf volume model} is then:
\begin{align}
E^{volume}(X,T,z;I) &= E^{u+p}(X) - w_3 f^{seg} (X,T,z) \nonumber \\ 
	& - w_4 |X \cap \pi| - w_5 |X^c \cap \pi^c|.
\label{eqn:volume-model}
\end{align}
where we call both $|X \cap \pi|$ and $|X^c \cap \pi^c|$, as the volume cost.
As an alternative to the sum of the intersections, we could have used the intersection over union metric. However, the sum of the intersections is easier to optimize, as we will see.

%%%%%%%%%%%%%%%%%%%%%%%%%%%%%%%%%%%%%%%%%%%%%%%%%%%%%%%%%%%%%%%%%%%%
% INFERENCE
\section{Pose Estimation and Image Segmentation}
\label{sec:inference}
Given the model and its parameters, we would like to estimate the image segmentation $X$ and 3D pose $T$. This is equivalent to solving the following optimization problem
\vspace{-1mm}
\begin{equation}
	(\hat{X},\hat{T}) = \operatorname*{argmin}_{X,T} E(X,T,z;I),
\label{eqn:joint-inference}
\end{equation}
where we do not solve for $z$ as we restrict ourselves to cars. In principle, this is a challenging mixed continuous and discrete optimization problem, where the energy is non-convex and non-smooth with respect the pose variables. Nonetheless, we show that this problem can be tackled using alternating minimization, where we estimate 3D pose given current segmentation, and then estimate segmentation given 3D pose till we reach convergence, \ie,
\begin{align}
	T^{n+1} &= \operatorname*{argmin}_T E(X^n, T, z; I) \label{eqn:pose-given-seg} \\
	X^{n+1} &= \operatorname*{argmin}_X E(X,T^{n+1},z;I). \label{eqn:seg-given-pose} 
\end{align}
We now discuss how to solve each sub-problem
%: segmentation given pose and pose given segmentation 
for the two models (shape model and volume model).

\subsection{Segmentation Given Pose}
\label{sec:inference-seg_given_pose}
The approach we use to solve Eqn.~\eqref{eqn:seg-given-pose} is to reformulate the energy into standard unary and pairwise terms, so that we can minimize it efficiently using graph cuts \cite{Boykov:PAMI01,Boykov:PAMI04}. More specifically, we show that the shape cost and volume costs can be decomposed into standard unary and pairwise terms.

\myparagraph{Shape cost} Let $r=(u,v,w)$ denote an arbitrary HOG bin, where $(u,v)$ denotes the pixel coordinates and $w$ is the orientation bin. Let $s$ be the superpixel corresponding to pixel $(u,v)$ and recall that $r^m(T,z)$ is the HOG bin associated with the $m$-th primitive projected under pose $T$. 
We can rewrite Eqn.~\eqref{eqn:modified-det-cost} in terms of
%Therefore, we can rewrite Eqn.~\eqref{eqn:modified-det-cost} using $(u,v,w)$ as
%Starting from Eqn.~\eqref{eqn:modified-det-cost}, we rewrite the equation in terms of 
%pixel indices $(u,v)$ and an orientation HOG index $w$ 
$(u,v,w)$ to get Eqn.~\eqref{eqn:fseg-decomp1}. 
%$r^m(T,z)$ is the pixel onto which the m-th edge primitive of model $z$ is projected to under pose $T$. $\bar{\mathcal{H}}(u,v,w)$ indicates that we read the HOG feature corresponding to pixel indices $(u,v)$ and orientation index $w$. 
Summing over $m$ and $w$ first, we get Eqn.~\eqref{eqn:fseg-decomp2} and then the sum over all pixels in the image is rewritten using superpixels \{$S_i$\}. Since, we assign the same label to all pixels inside a superpixel, we replace the $\delta(x_{s}=z)$ term with $\delta(x_i=z)$ for the i-th superpixel. This shows that $f^{seg}$ for a given pose $T$ can be written down as a 
%pairwise term between the image segmentation $X$ and the model class $z$ 
unary term as per Eqn.~\eqref{eqn:fseg-decomp-final}.
\vspace{-1mm}
\begin{align}
&\!\!\!
 f^{seg}(X,T,z;I) = \sum_{m=1}^{M_z} \bar{\mathcal{H}}(r^m(T,z)) \delta(x_{s^m(T,z)}=z) \nonumber \\
&\!\!\! = \!\sum_{m=1}^{M_z} \!\sum_{u,v,w} \!\! \bar{\mathcal{H}}(u,v,w) \delta((u,v,w) \!=\! r^m(T,z)) \delta(x_{s} \!=\! z) \!\!\label{eqn:fseg-decomp1}\\
&\!\!\! =\! \sum_{u,v} \mathcal{H}_1(u,v) \delta(x_{s}=z) \label{eqn:fseg-decomp2} \\
&\!\!\! =\! \sum_{i \in \mathcal{V}} \sum_{(u,v) \in S_i} \!\!\! \mathcal{H}_1 (u,v) \delta(x_{s} \!=\! z) 
 \!=\! \sum_{i \in \mathcal{V}} \mathcal{H}_2 (i) \delta(x_i \!=\! z).\!\!\! \label{eqn:fseg-decomp-final}
\end{align}

\myparagraph{Volume cost} For the volume cost, we have the terms $| X \cap \pi|$ and $|X^c \cap \pi^c|$. These are easily decomposed into pairwise terms between $X$ and $z$ as per the following equations:
\vspace{-2mm}
\begin{align}
|X \cap \pi(T,z)| &= \sum_{i \in \mathcal{V}} |S_i \cap \pi(T,z)| \delta(x_i = z), \label{eqn:acc1-decomp} \\
|X^c \cap \pi^c(T,z)| &= \sum_{i \in \mathcal{V}} |S_i \cap \pi^c(T,z)| \delta(x_i \neq z). \label{eqn:acc2-decomp}
\end{align}

\subsection{Pose Given Segmentation}
\label{sec:inference-pose_given_seg}

As we will show below, we can reformulate the pose-given-segmentation sub-problem in Eqn.~\eqref{eqn:pose-given-seg} as a maximization problem of the form: $\hat{T} = \operatorname*{argmax}_{T \in \mathbb{T}} f(T,X,z;I)$ with corresponding score $f$. This is a highly non-convex problem with a large search space. Prior work \cite{Yoruk:3DRR13} used a B\&B strategy to address the pose estimation without any segmentation information, which involves maximizing $f = f^{det}$ in Eqn.~\eqref{eqn:original-det-cost}. More generally, B\&B algorithms have been used extensively in the computer vision literature \cite {Lampert:PAMI09, Kokkinos:NIPS11, Schwing:ECCV12, Sun:AISTATS12} to obtain the global maxima of non-convex problems in an efficient manner. The key idea behind B\&B methods is to start with the whole search space and successively refine it while maintaining a priority queue of candidate solutions till the desired resolution is achieved. Each solution corresponds to a region in the search space, which we refer to as a pose cell. The queue is maintained using the upper bound of the score in each cell. This works because a pose cell containing the global maxima is expected to have a higher upper-bound compared to other cells. We refer the reader to \cite{Yoruk:3DRR13} for more details on how B\&B is applied to the pose estimation task without any segmentation information. The key takeaway for us is that we can find the global maxima efficiently as long as we can find an upper bound $\bar{f}$ for any given pose cell $\mathbb{T}$ in the search space. This corresponds to Eqn.~\eqref{eqn:upper-bound}. In our case, the pose $T$ consists of the azimuth $a$, the elevation $b$, the camera-tilt $c$, the depth $d$ and a 2D translation in the image plane $(u,v)$. A corresponding pose cell $\mathbb{T}$ is then of the form $\mathbb{T} = [a_1,a_2] \times [b_1,b_2] \times [c_1,c_2] \times [d_1,d_2] \times [u_1,u_2] \times [v_1,v_2]$.
\begin{equation}
\!\!\!\bar{f} \geq \max_{T \in \mathbb{T}} f(T) =\!\! \max_{(a,b,c,d,u,v) \in \mathbb{T}} f(T\!=\!(a,b,c,d,u,v)).\! \!
	\label{eqn:upper-bound}
\end{equation}

\begin{figure}[b]
	\centering
	\begin{subfigure}[b]{0.45\linewidth}
		\includegraphics[width=\linewidth]{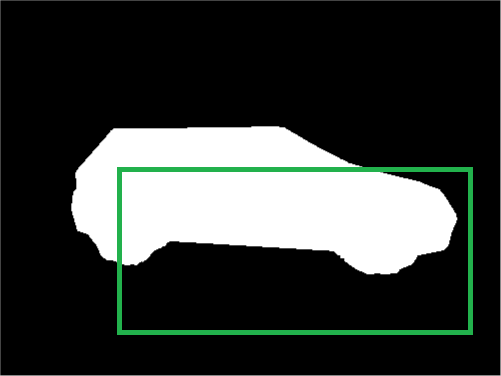}	
		\caption{$X$}
		\label{fig:ub1_seg}
	\end{subfigure}
	~
	\begin{subfigure}[b]{0.45\linewidth}
		\includegraphics[width=\linewidth]{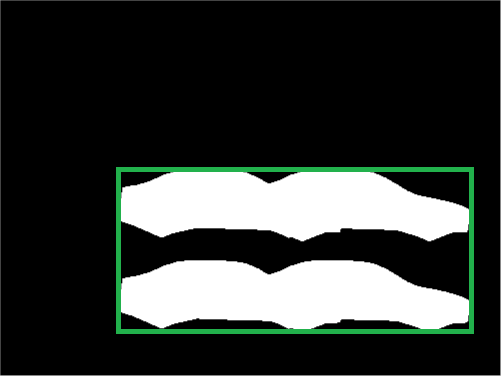}
		\caption{combined mask}
		\label{fig:ub1_mask}
	\end{subfigure}
	\caption{Computing upper bound of $|X \cap \pi(T)|$.}
	\label{fig:ub1}
\end{figure}

\myparagraph{Shape model}
Solving Eqn.~\eqref{eqn:pose-given-seg} for the shape model is equivalent to solving the following problem:
\begin{align}
\hat{T} &= \operatorname*{argmin}_T E^{shape}(X,T,z; I) \nonumber \\
	&= \operatorname*{argmin}_T E^{u+p}(X;I) - w_3 f^{seg} (X,T,Z) \nonumber \\
	&= \operatorname*{argmax}_T f^{seg}(X,T,Z). \label{eqn:shape-pose-given-seg}
\end{align}
This problem is identical to the one solved in \cite{Yoruk:3DRR13} except that now we have the HOG features extracted only from the foreground because of the $\delta(x_{s^m(T)}=z)$ term in Eqn.~\eqref{eqn:modified-det-cost}.

\myparagraph{Volume model}
Solving Eqn.~\eqref{eqn:pose-given-seg} for the volume model gives us Eqn.~\eqref{eqn:volume-pose-given-seg} where $\lambda_1 = \frac{w_4}{w_3}$ and $\lambda_2= \frac{w_5}{w_3}$.
\begin{align}
\hat{T} &= \operatorname*{argmin}_T E^{volume}(X,T,z; I) \nonumber \\
	&= \operatorname*{argmin}_T E^{u+p}(X;I) - w_3 f^{seg}(X,T,z) \nonumber \\
	& \hspace{4em}- w_4 |X \cap \pi(T,z)| - w_5 |X^c \cap \pi^c(T,z)| \nonumber \\
	&= \operatorname*{argmax}_T f^{seg}(X,T,Z) + \lambda_1 |X \cap \pi(T,z)| \nonumber \\ 
	&\hspace{4em} + \lambda_2 |X^c \cap \pi^c(T,z)| \label{eqn:volume-pose-given-seg}
\end{align}
We can define the upper bound $\bar{f^{vol}}$ as the sum of bounds over each of the three terms individually, \ie,
\begin{align}
	 \bar{f^{vol}} & \geq \max_{T \in \mathbb{T}} f^{seg}(T) + \lambda_1 \max_{T \in \mathbb{T}} |X \cap \pi(T)| \nonumber \\ 
	&+ \lambda_2 \max_{T \in \mathbb{T}} |X^c \cap \pi^c(T)|.
\end{align}
We now discuss how to find bounds for each term. 

\mysubparagraph{First term}
The first term is identical to the one in the shape model and its upper bound is computed the same way. 

\mysubparagraph{Second term}
%\underline{Bounding $\max_{T \in \mathbb{T}} |X \cap \pi(T)|$}: 
Given a cell $\mathbb{T} = [a_1,a_2] \times [b_1,b_2] \times [c_1,c_2] \times [d_1,d_2] \times [u_1,u_2] \times [v_1,v_2]$, we want to find an upper bound for the overlap $|X \cap \pi(T)|$ between the segmentation, $X$, and the projection of the car, $\pi(T)$. Higher overlap is obtained if the projected car is bigger, so we choose the parameters $(\hat{a},\hat{b},\hat{c},\hat{d})$ to maximize the projection of the car. The car is bigger when viewed from the side, so we choose $\hat{a}=0$, $-180$ or $180$ if it lies in $[a_1, a_2]$, or else $\hat{a} = a_1, a_2$, whichever is closest to $0$, $-180$ or $180$. A larger car is observed also from a higher elevation, so we choose $\hat{b} = \max(b_1,b_2)$. A bigger car is observed also when it is closer to the camera, so we choose $\hat{d}=\min(d_1,d_2)$. The camera tilt $c$ does not affect car size and is chosen as $\hat{c}=\min(c_1,c_2)$. Once, we obtain $(\hat{a},\hat{b},\hat{c},\hat{d})$, we project it onto the image at $(u,v) = [(u_1,v_1), (u_1,v_2), (u_2,v_1), (u_2,v_2)]$ to get four masks. These are combined together to get Figure \ref{fig:ub1_mask}. The green rectangle bounds the spatial extent in which the projection might lie in pose cell $\mathbb{T}$. 

\begin{figure}[h]
	\centering
	\begin{subfigure}[b]{0.45\linewidth}
		\includegraphics[width=\linewidth]{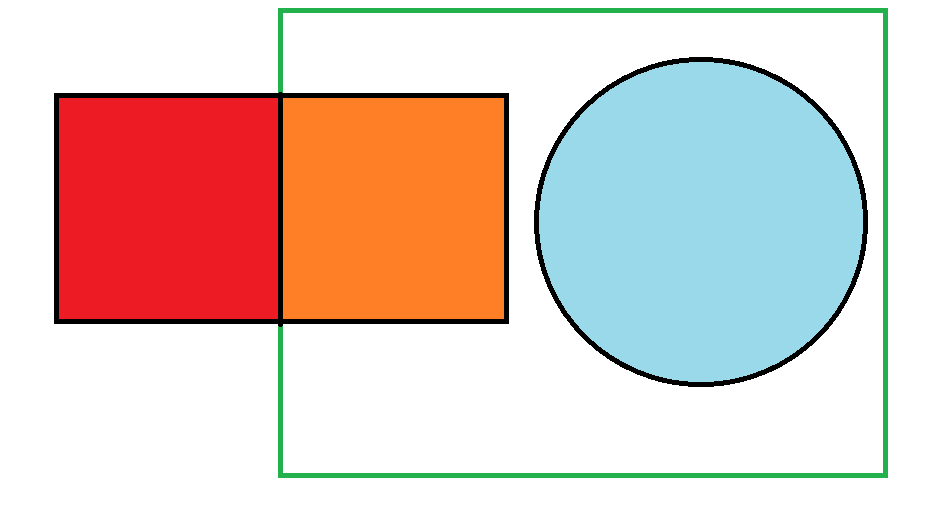}
	\end{subfigure}
	~
	\begin{subfigure}[b]{0.45\linewidth}
		\includegraphics[width=\linewidth]{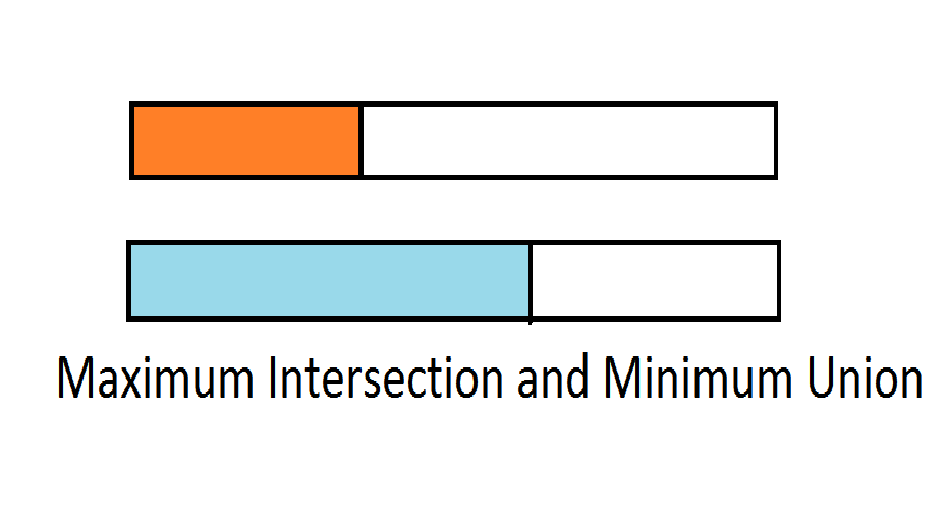}
	\end{subfigure}
	\caption{Bound computation (best seen in color).}
	\label{fig:ub_comp}
\end{figure}

Now, the problem is to find a bound of the overlap allowing all possible projections inside the green rectangle. This is the same as finding an overlap between the rectangle and the circle in Figure \ref{fig:ub_comp}, where the circle is allowed to lie anywhere inside the green boundary. Let $s_1$ be the number of pixels in the rectangle inside the boundary (shown in orange color) and $s_2$ be the number of pixels in the circle. An upper bound on the intersection $|orange \cap blue|$ is obtained as the minimum of these two numbers. This is depicted in the two rectangles with orange and blue bars indicating the number of foreground pixels of each figure inside the green boundary. We use the green rectangle and the segmentation $X$ (Figure \ref{fig:ub1_seg}) to get score $s_1$.
% as the number of foreground pixels inside the red rectangle. 
Choosing $(u,v)=(0,0)$ and projecting onto the image gives us a mask of the largest projected car. $s_2$ is the number of foreground pixels in this mask. The desired bound on $|X \cap \pi(T)|$ is $\min(s_1,s_2)$.

\vspace{-3mm}

\begin{figure}[h]
	\centering
	\begin{subfigure}[b]{0.45\linewidth}
		\includegraphics[width=\linewidth]{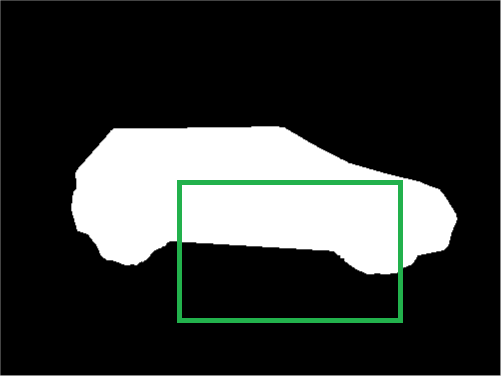}
		\caption{X}
		\label{fig:ub2_seg}
	\end{subfigure}
	~
	\begin{subfigure}[b]{0.45\linewidth}
		\includegraphics[width=\linewidth]{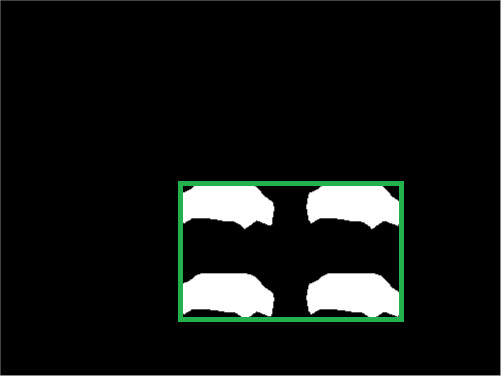}
		\caption{combined mask}
		\label{fig:ub2_mask}
	\end{subfigure}
	\caption{Computing lower bound of $|X \cup \pi(T)|$.}
	\label{fig:ub2}
\end{figure}

\mysubparagraph{Third term}
%\underline{Bounding $\max_{T \in \mathbb{T}} |X^c \cap \pi^c(T)|$ } : 
Note that an upper bound for this term can be obtained from a lower bound for the union term $|X \cup \pi(T)|$:
\begin{align}
	\max_{T \in \mathbb{T}} |X^c \cap \pi^c(T)| &= \max_{T \in \mathbb{T}} \left ( |I| - |X \cup \pi(T)| \right ) \nonumber \\
				&= |I| - \min_{T \in \mathbb{T}} |X \cup \pi(T)|.
\end{align}
To minimize the union between $X$ and $\pi(T)$ we need the smallest possible projection of the car. This is obtained by choosing the parameter $\tilde{a} = -90,90$ if it lies in $[a_1,a_2]$ or choosing from $a_1,a_2$ whichever is closest to one of $-90,90$. $\tilde{b}=\min(b_1,b_2)$, $\tilde{c}=\min(c_1,c_2)$ and $\tilde{d}=\max(d_1,d_2)$ are chosen to get the smallest projection. A similar procedure to the one described for the intersection term is followed (shown in Figure \ref{fig:ub2}). The difference now is that we want to minimize the union. Let $s_3$ be the number of foreground pixels outside the green rectangle, $s_4$ be the number of foreground pixels inside the blue rectangle in segmentation $X$, and $s_5$ be the number of foreground pixels in the projected model at $(\tilde{a},\tilde{b},\tilde{c},\tilde{d},0,0)$. The smallest union is obtained when one lies completely inside the other and is given by $\max(s_4, s_5)$ to give the lower bound on the union as $s_3 + \max(s_4,s_5)$. The corresponding upper bound on $|X^c \cup \pi^c(T)|$ is $|I| - [s_3 + \max(s_4,s_5)]$.

\vspace{-2mm}
\vspace{-4mm}

% {\color{red} Explain here issue about only finding best cell, that best pose is chosen as center of the cell, but that this may affect the convergence, and how you fix it by using cells of multiple resolutions. Also, initialization?}
\myparagraph{Bound Analysis} The upper bounds discussed here are based on the simple inequalities $|A \cap B| \leq \min(|A|,|B|)$ and $|A \cup B| \geq \max(|A|,|B|)$. For an empirical analysis of how tight these bounds actually are (short answer: pretty tight), we refer the reader to supplementary material.

%%%%%%%%%%%%%%%%%%%%%%%%%%%%%%%%%%%%%%%%%%%%%%%%%%%%%%%%%%%%%%%%%%%%
% LEARNING
\section{Learning the model parameters}
\label{sec:learning}

We need to learn the parameters of the energy functions $E^{shape}$ and $E^{volume}$. In our case, the energy functions are linear in their parameters $w$, \ie, $E^{shape} = w_{shape}^T \Psi^{shape}$ and $E^{volume} = w_{volume}^T \Psi^{volume}$. We follow the max-margin approach to learn the parameters as: 
%
%\begin{equation}
%\label{eqn:learning}
%\begin{split}
%w = \operatorname*{argmin}_w \frac{\lambda}{2}\|w\|_2^2 + \frac{1}{N} \sum_{i=1}^N \xi_i ~~\text{s.t.} ~~ w \geq 0 ~~\text{and}~~  \forall (i, X_i, T_i), \\    E(X_i,T_i) - E(X_i^*,T_i^*) \geq \Delta(X_i,X_i^*,T_i,T_i^*) - \xi_i
%\end{split}
%\end{equation}
\vspace{-2mm}
\begin{align}
\label{eqn:learning}
& w = \operatorname*{argmin}_{w\geq 0} \frac{\lambda}{2}\|w\|_2^2 + \frac{1}{N} \sum_{i=1}^N \xi_i ~~\text{s.t.} ~~ \forall (i, X_i, T_i), \nonumber \\   
& E(X_i,T_i) - E(X_i^*,T_i^*) \geq \Delta(X_i,X_i^*,T_i,T_i^*) - \xi_i.\!
\end{align}
Here, $X_i^*$ and $T_i^*$ are the ground truth segmentation and ground truth pose of image $i$ respectively, $X_i$ is any possible segmentation and $T_i$ is any possible pose. The optimization problem in Eqn.~\eqref{eqn:learning} is similar to the Structural SVM problem \cite{Tsochantaridis:JMLR05, Joachims:JMLR09} and we solve it using the Block Coordinate Frank Wolfe algorithm \cite{Julien:JMLR13}. The only difference from the standard Structural SVM problem is that we need non-negative weights. This is to keep the energy function sub-modular (for efficient inference using graph cuts). Non-negativity is enforced by adding a thresholding step in the algorithm described in \cite{Julien:JMLR13}. Now, since the number of constrains in Eqn.~\eqref{eqn:learning} is exponentially large, it is customary to solve the problem by adding one constraint at a time. We find the most violated constraint by solving a loss-augmented inference problem which is identical to the optimization problems shown in \S\ref{sec:inference-seg_given_pose} except that we now have additional unary terms introduced by the loss function $\Delta(X,X^*,T,T^*)=\Delta_1(X,X^*)+\Delta_2(T,T^*)$. We use the standard hamming loss for the segmentations and azimuth error loss ($\frac{\min( |a-a^*| ,360^\circ-|a-a^*| )}{360^\circ}$) for the pose.
% 

%%%%%%%%%%%%%%%%%%%%%%%%%%%%%%%%%%%%%%%%%%%%%%%%%%%%%%%%%%%%%%%%%%%%
% RESULTS
\section{Results}
\label{sec:results}
We report results on three sets of experiments to demonstrate the power of our proposed models:
\begin{enumerate}[topsep=2pt,itemsep=-1ex,partopsep=1ex,parsep=1ex]
	\item Estimate segmentation given ground truth pose.
	\item Estimate pose given ground truth segmentation.
	\item Estimate both pose and segmentation using the alternating minimization (AM) procedure in Eqns.~\eqref{eqn:pose-given-seg}-\eqref{eqn:seg-given-pose}.
\end{enumerate}
The first two sets of experiments aim to verify our hypothesis: knowledge of segmentation can help improve pose estimation and knowledge of pose can help improve segmentation, while the third one applies our models to data without any ground truth information and evaluates performance.

%--------------------------------------------- Dataset
\myparagraph{Datasets}
\label{sec:results-dataset}
We fix object class label $z$ to the class `car'. Therefore, the problem is to segment the image into car regions (foreground) and non-car regions (background). 
%We used the wireframe model of a car learnt by \cite{Yoruk:3DRR13}. 
We use the Pascal VOC dataset, which has annotations for both image segmentations and 3D fine pose. Pose annotations for Pascal VOC2011 images were provided as part of the Pascal 3D+ dataset \cite{Xiang:WACV14}. Image segmentations were obtained from the Semantic Boundary dataset \cite{Hariharan:ICCV2011}, which annotated all Pascal VOC2012 trainval images. We use the splits recommended for the Pascal VOC2011 challenge and use the val data for testing. Restricting ourselves to just car images that have one car, we have 670 images in total with 504 images for training and 166 images for testing. We also test the pose estimation given ground-truth segmentation experiments on 480 car images of the 3DObject dataset \cite{Savarese-FeiFei:ICCV07}.

\myparagraph{Object model} We use the car wireframe model  $\mathcal{M}$ from \cite{Yoruk:3DRR13}, which was trained using 10 blueprints of sedan cars.

\myparagraph{Projection} We use alpha-shapes with the circumradius as the $95^{th}$ percentile of the edges in the Delaunay triangulation to generate $\pi(T)$. This is closer to how a car looks compared to a convex-hull, which discards the car's shape.

\myparagraph{Superpixel Graph} Superpixels were generated by the Quickshift algorithm \cite{ComaniciuD2002-PAMI, Vedaldi:ECCV08-qs} using the VL-feat library \cite{vedaldi:vlfeat}.

\myparagraph{Bottom-up potentials} The unaries $\phi_i(x_i;I)$ are learnt using Textonboost \cite{Shotton:IJCV09} (code provided by densecrf \cite{Krahenbuhl:NIPS11}). We do not use the DPM detector output, as recommended, to generate purely bottom up unaries. A standard pairwise potential of the form $\psi_{ij}(x_i,x_j;I) = \frac{L_{ij}}{1+\gamma \|I_i - I_j\|} \delta(x_i \neq x_j)$ is used. Here, $L_{ij}$ is the length of the common boundary between superpixels $i$ and $j$. $I_i$ and $I_j$ are the mean intensities of the superpixels in the LUV color space.

%---------------------------------------------Segmentation given gt-pose
\subsection{Segmentation Given Ground Truth Pose}
\label{sec:results-seg_given_gt_pose}
We assume we know the ground truth 3D pose and estimate the segmentation using our two models. We compare with some standard baselines generated using the Joint Categorization and Segmentation framework \cite{Singaraju:CVPR11} with their unary+pairwise and unary+pairwise+linear-top-down models. These baselines are all image-based as they do not include any kind of 3D pose information. We also report some results from the Pascal 3D+ dataset paper \cite{Xiang:WACV14} for comparison. \cite{Xiang:WACV14} used CAD models of 10 kinds of cars to annotate the 3D pose of cars in the image. They  reprojected the CAD model onto the image at the annotated pose and generated a segmentation. This is an upper bound on the best segmentation performance we can achieve (shown in red in Table~\ref{table:seg-given-gt-pose}). For the Viewpoint DPM (VDPM-16) results, they fixed the depth, elevation, camera-tilt and image-translation parameters to their ground truth values and  estimated the azimuth using a variation of DPMs. Then, they projected the CAD model at the estimated pose to generate a segmentation. We generate one more baseline by projecting the wireframe model of \cite{Yoruk:3DRR13} onto the image using the ground truth pose and generate a candidate segmentation.  %Finally, we find segmentations using the shape and volume models and compare results.

\myparagraph{Learning} Model weights are learnt using Eqn.~\eqref{eqn:learning} with constraints $E(X_i,T_i^*)-E(X_i^*,T_i^*) \geq \Delta_1(X_i,X_i^*)-\xi_i$.

\begin{table}[h]
	\centering
	\begin{tabular}{|@{\,}c@{\,}|@{\,\,}c@{\,\,}|@{\,\,}c@{\,\,}|@{\,\,}c@{\,\,}|}
	\hline
		Method & Bg. & Cars & Mean \\
	\hline
		Unary \cite{Singaraju:CVPR11} & 87.53 & 51.51 & 69.52 \\
		Unary+Pairwise \cite{Singaraju:CVPR11}& 89.17 & 54.60 & 71.88 \\
		U+P+Linear-Top-down \cite{Singaraju:CVPR11}& 88.05 & 54.54 & 71.29 \\
	\hline
		Projected CAD Model \cite{Xiang:WACV14} & - & {\color{red}67.30} & - \\
		VDPM-16 \cite{Xiang:WACV14} & - & 51.9 & - \\
		Projected Wireframe Model \cite{Yoruk:3DRR13} & 91.04 & 61.90 & 76.47 \\
	\hline
		Shape model (ours) & 89.96 & 59.08 & 74.52 \\
		Volume model (ours) & {\bf 91.46} & {\bf 64.60} & {\bf 78.03} \\
	\hline
	\end{tabular}
	\caption{Segmentation results (Intersection over Union - I/U score) for segmentation-given-ground-truth-pose experiments. The best results are in bold. See text for more details.}
	\label{table:seg-given-gt-pose}
\end{table}

As can be seen in Table~\ref{table:seg-given-gt-pose}, using only image information we get upto 54.60 I/U score for the foreground (class cars). Projecting the wireframe model onto the image and not using any image information gives us pretty good results (61.90 I/U score). However, the best results are obtained for the volume model at 64.60 I/U score by combining both the bottom-up information contained in unary and pairwise terms and the top-down information contained in the 3D object models. This experiment also demonstrates that while the shape model does better than pure bottom-up models, it needs the volume cost to compete with the projection baselines. Overall, our result of 64.60 I/U is the closest to the best possible achievable segmentation obtained by projecting the true 3D model with the true 3D pose.

%--------------------------------------------Pose given gt-seg
\subsection{Pose Given Ground Truth Segmentation}
\label{sec:results-pose_given_gt_seg}
In this set of experiments, we estimate the 3D pose of the object present in the image given the ground truth segmentation. We use the parameter ranges for the pose parameters as mentioned in Table~\ref{table:pose-ranges}. Elevation and depth are measured from the center of mass of the car model. Image-plane translation $(u,v)$ is in degrees assuming the width of the image corresponds to a field of view of $55^\circ$.

\myparagraph{Learning} Model weights are learnt using Eqn.~\eqref{eqn:learning} with constraints $E(X_i^*,T_i)-E(X_i^*,T_i^*) \geq \Delta_2(T_i,T_i^*)-\xi_i$.

\begin{table}[h]
	\centering
	\begin{tabular}{|c|c|c|}
	\hline
		Parameter & Range & Resolution \\
	\hline
		Azimuth & $[-180^{\circ}, 180^{\circ}]$ & $5.62^{\circ}$ \\
		Elevation & [0.0m, 3.0m] & 0.75 m \\
		Camera-Tilt & $[0^{\circ}, 0^{\circ}]$ & 0 \\
		Depth & [4m, 50m] & 1.44m \\
		Image-Trans & $[-27.5^\circ,27.5^\circ]$ & $3.44^\circ$ \\
	\hline
	\end{tabular}
	\caption{\label{table:pose-ranges}Parameter ranges and highest resolution passed to the branch and bound algorithm for pose estimation}
\end{table}

We also do a pre-processing step in some experiments (indicated by -DPM) to restrict the pose ranges based on the output of a `car' DPM detector trained on ImageNet data. Given detected bounding box, we check over ranges of depth to restrict the search to only those depths where the projected model has some overlap with the bounding box ($>0.5$ intersection over union score). We also restrict the search over image translations $(u,v)$ to be $\pm5^\circ$ around the center of the detected bounding box.

\begin{figure}
	\centering
	\begin{subfigure}[b]{0.48\linewidth}
		\includegraphics[width=\linewidth]{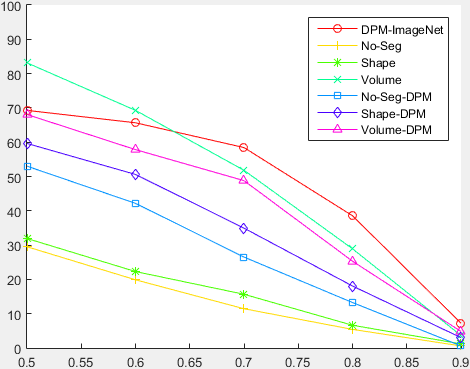}
		\caption{Detection accuracy}
		\label{fig:pose-given-gtseg-voc2011:acc}
	\end{subfigure}
	~
	\begin{subfigure}[b]{0.48\linewidth}
		\includegraphics[width=\linewidth]{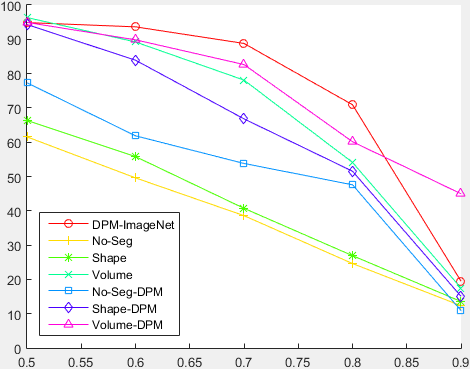}
		\caption{Average Precision}
		\label{fig:pose-given-gtseg-voc2011:ap}
	\end{subfigure}
\caption{Evaluation of pose estimation given ground-truth segmentation using 2D object detection metrics}
\label{fig:pose-given-gtseg-voc2011}
%\end{figure}
\smallskip
%\begin{figure}[h]
	\centering
	\begin{subfigure}[b]{0.48\linewidth}
	\includegraphics[width=\linewidth]{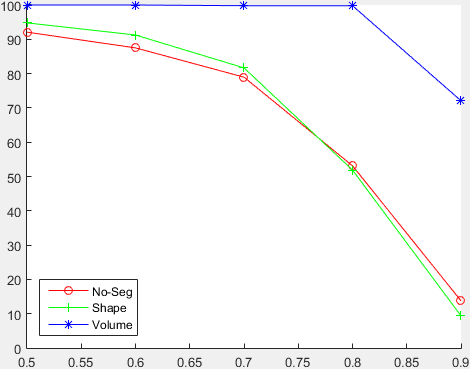}
	\caption{Detection Accuracy}
	\label{fig:pose-given-gtseg-3dobject:acc}
	\end{subfigure}
	~
	\begin{subfigure}[b]{0.48\linewidth}
	\includegraphics[width=\linewidth]{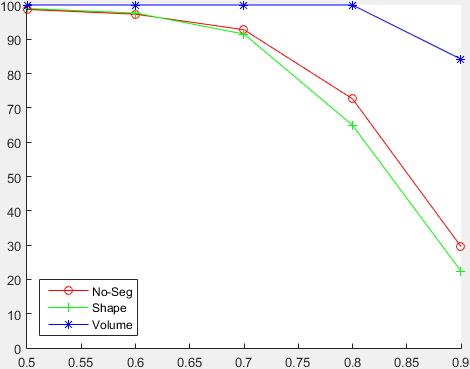}
	\caption{Average Precision}
	\label{fig:pose-given-gtseg-3dobject:ap}
	\end{subfigure}
\caption{Object detection results for 3Dobject dataset. Weights for the volume model are same as learnt from VOC data}
\label{fig:pose-given-gtseg-3dobject}
\end{figure}

Figure \ref{fig:pose-given-gtseg-voc2011} shows 2D object detection results of the 3D pose estimation task, where the model is projected onto the image at estimated 3D pose to get a 2D bounding box detection. We show detection accuracy (percentage of images where the car was detected correctly) and average precision score for different models and detection thresholds (I/U score between detected and ground-truth bounding boxes). 

As can be seen in Figure \ref{fig:pose-given-gtseg-voc2011:acc}, the pose estimation task done without any segmentation (yellow) gives the worst results. Performance slightly improves if we use the shape model (green). A noticeable improvement is observed when we restrict the pose search-space using the output of a DPM for both the no-segmentation (blue) and shape models (violet). However, the best performance is observed by the volume model, both without any restriction in pose-space (cyan) and with restricted pose-space (pink). The DPM output is shown in red. Observe that DPM performs worse at the typical 0.5 I/U threshold, and better at higher thresholds. Similar trends are observed in Figure~\ref{fig:pose-given-gtseg-voc2011:ap}. Overall, using a simple 3D model and well-designed scores, we achieve a performance on par with a highly specialized 2D detection system trained on lots of images.
%We are able to achieve performance on par with DPM, a complicated 2D detection system trained over lots of images for a long time, using a simple 3D model and well-designed scores. 
In Figure \ref{fig:pose-given-gtseg-3dobject}, we test the shape and volume models on the 3Dobject dataset. The best performance is observed for the volume model with comparable performance for the shape and no-seg models. 

The 3Dobject dataset has 480 car images, with 10 cars shown at 8 azimuths, 2 elevations and 3 depths each. Performance of 3D pose estimation can be evaluated by computing the azimuth classification accuracy (into 8 discrete labels/directions). As can be seen in Table~\ref{table:pose-given-gtseg-3dobject}, the volume model performs the best. For the case of VOC data where we have fine 3D pose annotations, we compute oriented detection accuracy, which is similar to standard object-detection metrics, except that we call a detection valid if it has significant I/U overlap ($>0.5$) with the ground-truth bounding box and an azimuth error (estimated azimuth - ground-truth azimuth) below a threshold ($20^\circ$). As can be seen from  Table~\ref{table:pose-given-gtseg-voc2011:oriented}, similar trends are observed as in the object detection results. 

\begin{table}[h]
\centering
\begin{tabular}{|c|c|c|c|}
\hline
Model & No-Seg & Shape & Volume \\
\hline
Accuracy & 78.13 & 85.42 & 91.67 \\
\hline
\end{tabular}
\caption{Classification accuracy for azimuth on 3Dobject dataset}
\label{table:pose-given-gtseg-3dobject}
%\end{table}
\medskip
%\begin{table}[h]
\centering
\begin{tabular}{|c|c||c|c|}
\hline
Method & Accuracy & Method & Accuracy \\
\hline
No-Seg & 11.45 & No-Seg-DPM & 19.28 \\
Shape & 13.86 & Shape-DPM & 23.49 \\
Volume & 27.11 & Volume-DPM & 28.31 \\
\hline
\end{tabular}
\caption{Oriented detection accuracy for pose estimation experiments on VOC2011 data using ground-truth segmentation}
\label{table:pose-given-gtseg-voc2011:oriented}
\end{table}

%-------------------------------------------Joint pose and seg
\subsection{Joint Segmentation and Pose Estimation Using Alternating Minimization}
\label{sec:results-joint_pose_and_seg}
We now present the results of pose estimation and segmentation using the AM procedure in Eqns.~\eqref{eqn:pose-given-seg}-\eqref{eqn:seg-given-pose}. We initialize $X^0$ as the output of the Unary+Pairwise problem (purely bottom-up segmentation) and $T^0$ as the output of the No-Segmentation-DPM problem (3D pose estimation without any ground-truth segmentation with pose search-space restricted by DPM output). At every step of the AM procedure we require that $E(X^{n+1},T^{n+1}) \leq E(X^n,T^n)$. Note that we require point estimates of pose $T^n$ whereas B\&B gives us a pose cell. This difficulty is overcome by maintaing a best pose estimate during the B\&B procedure. This is given as $(\hat{X}=\operatorname*{argmin}_X E(X,\hat{T}), \hat{T})$ where $\hat{T}$ is the center of the pose cell being evaluated in the B\&B queue. We refine the pose search-space till we obtain $E(\hat{X},\hat{T}) < \tau E(X^n,T^n)$ for some suitably chosen threshold parameter $\tau$. The AM procedure is said to converge when segmentation doesn't change, \ie $\frac{|X^n \cap X^{n+1}|}{|X^n \cup X^{n+1}|} > 0.98$.

\begin{table}[h]
\centering
\begin{tabular}{|@{\;}c@{\;}|c|c|c|c|@{\;}c@{\;}|}
\hline
Method & \multicolumn{3}{|c|}{Segmentation} & \multicolumn{2}{|c|}{Detection} \\
\hline
& Bg. & Cars & Mean & Obj. & Oriented \\
\hline
B1-Shape & {\bf 89.17} & {\bf 54.60} & {\bf 71.88} & 45.78 & 18.67 \\
B1-Volume & 89.17 & 54.60 & 71.88 & 43.98 & 10.24 \\
\hline
B2 -Shape & 87.97 & 54.19 & 71.08 & 53.01 & 19.28 \\
B2 -Volume & 83.52 & 45.06 & 64.29 & 53.01 & 19.28 \\
\hline
AM-Shape & 87.69 & 54.29 & 70.99 & 53.01 & 21.69 \\
AM-Volume & 86.42 & 50.75 & 68.58 & {\bf 56.02} & {\bf 22.29} \\
\hline
\end{tabular}
\caption{Segmentation results (I/U score) and pose estimation results (object detection accuracy and oriented object detection accuracy) for the joint experiments}
\label{table:joint-results}
\end{table}

We compare our results with two baselines, $B1: (X,T)=(X^0, \operatorname*{argmin}_T E(X^0,T))$ and $B2: (X,T) = (\operatorname*{argmin}_X E(X,T^0),T^0)$. These are essentially the AM procedure stopped after 1 iteration. As can be seen in Table~\ref{table:joint-results}, for the Shape model, the AM procedure leads to similar segmentation and detection results but marginally better oriented detections. There is trade-off in the volume model, where it loses some segmentation accuracy for better pose estimation. This is to be expected as we are trying to describe complex images in the VOC data with significant intra-class shape variation, occlusion and truncation using a simple sedan car model. 

%%%%%%%%%%%%%%%%%%%%%%%%%%%%%%%%%%%%%%%%%%%%%%%%%%%%%%%%%%%%%%%%%%%%
% All results for one car
\begin{figure}
	\centering
	\begin{subfigure}[b]{0.45\linewidth}
		\includegraphics[width=\linewidth]{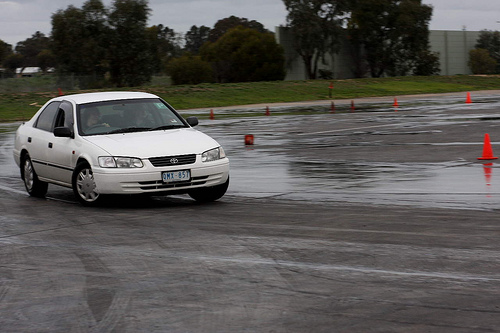}
		\caption{Image}
	\end{subfigure}
	~
	\begin{subfigure}[b]{0.45\linewidth}
		\includegraphics[width=\linewidth]{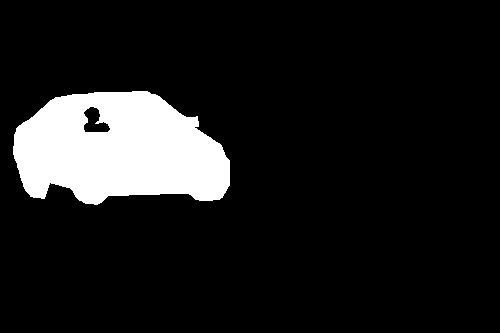}
		\caption{GT Segmentation}
	\end{subfigure}

	\begin{subfigure}[b]{0.45\linewidth}
		\includegraphics[width=\linewidth]{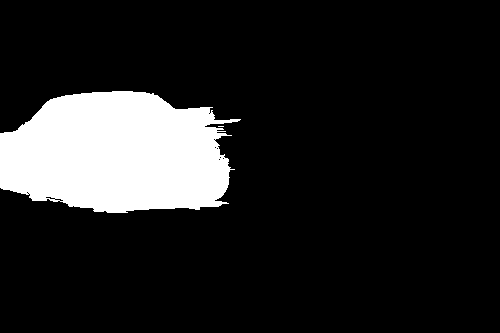}
		\caption{Segmentation using GT pose under shape model}
	\end{subfigure}
	~
	\begin{subfigure}[b]{0.45\linewidth}
		\includegraphics[width=\linewidth]{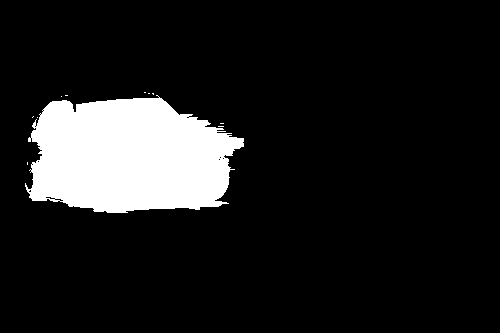}
		\caption{Segmentation using GT pose under volume model}
	\end{subfigure}

	\begin{subfigure}[b]{0.45\linewidth}
		\includegraphics[width=\linewidth]{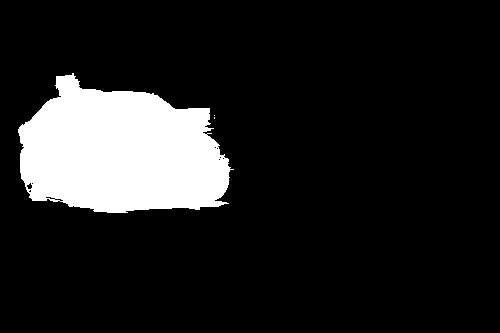}
		\caption{Segmentation after 5 iterations of AM under shape model}
	\end{subfigure}
	~
	\begin{subfigure}[b]{0.45\linewidth}
		\includegraphics[width=\linewidth]{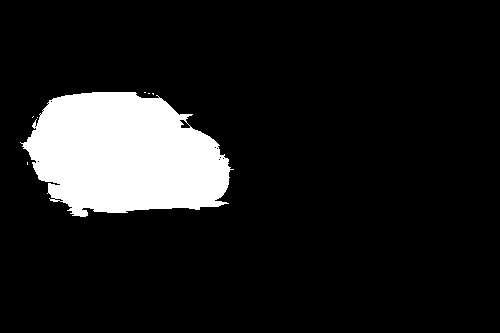}
		\caption{Segmentation after 8 iterations of AM under volume model}
	\end{subfigure}

	\begin{subfigure}[b]{0.45\linewidth}
		\includegraphics[width=\linewidth]{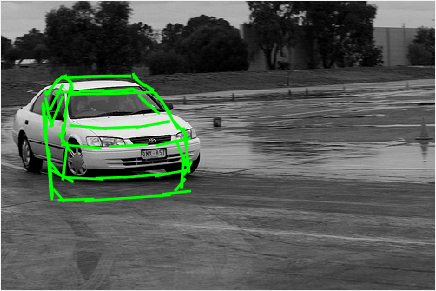}
		\caption{Pose using GT segmentation under shape model}
	\end{subfigure}
	~
	\begin{subfigure}[b]{0.45\linewidth}
		\includegraphics[width=\linewidth]{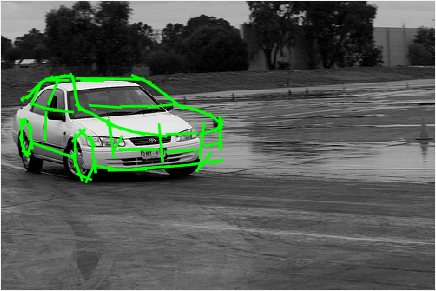}
		\caption{Pose using GT segmentation under volume model}
	\end{subfigure}

	\begin{subfigure}[b]{0.45\linewidth}
		\includegraphics[width=\linewidth]{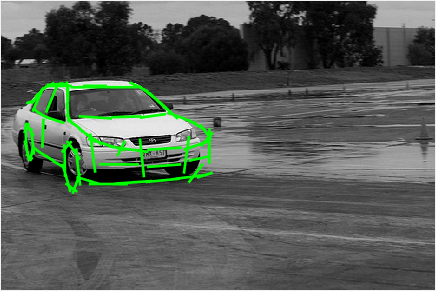}
		\caption{Pose estimate after 5 iterations of AM under shape model}
	\end{subfigure}
	~
	\begin{subfigure}[b]{0.45\linewidth}
		\includegraphics[width=\linewidth]{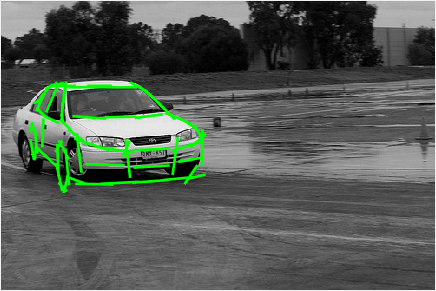}
		\caption{Pose estimate after 8 iterations of AM under volume model}
	\end{subfigure}
	\caption{The output of various experiments for one image}
	\vspace{-6mm}
\end{figure}

%%%%%%%%%%%%%%%%%%%%%%%%%%%%%%%%%%%%%%%%%%%%%%%%%%%%%%%%%%%%%%%%%%%%
% CONCLUSION
\section{Conclusion}
We introduced two new top-down costs based on 3D object models that capture the consistency between 3D shape and 2D appearance, and 3D volume and 2D segmentation. We demonstrated that these costs can be decomposed into standard unary and pairwise terms for inference using graph-cuts. We also computed  upper-bounds for these costs allowing efficient pose estimation using branch-and-bound algorithms. We demonstrated good performance on estimating segmentation given ground truth pose and estimating pose given ground truth segmentation. We also observed slight improvement in 3D pose estimation when solving for both pose and segmentation. However, this comes at the cost of reduced accuracy in segmentation.
Arguably this is because of the use of a simple rigid model for sedans, which is not able to capture the intra-class variability. Nonetheless, the overall results are quite promising, and will only improve with the availability of better 3D models.

\myparagraph{Acknowledgement} This work was supported by NSF grant 1527340.

{\small
\bibliographystyle{ieee}
\bibliography{biblio/vidal,biblio/vision,biblio/segmentation,biblio/recognition,biblio/learning}
}

\end{document}